%% file: main.tex
\pdfoutput=1

\documentclass[11pt]{article}

\usepackage[final]{acl}

\usepackage{times}
\usepackage{latexsym}
\usepackage{makecell}
\usepackage{tikz}
\usetikzlibrary{tikzmark}

\usepackage[T1]{fontenc}

\usepackage[utf8]{inputenc}

\usepackage{microtype}

\usepackage{inconsolata}

\usepackage{graphicx}

\usepackage{amsmath,amssymb,amsfonts}
\usepackage{algorithmic}
\usepackage{graphicx}
\usepackage{booktabs}
\usepackage{textcomp}
\usepackage{arydshln}
\usepackage{xcolor}
\usepackage{multirow}
\usepackage[most]{tcolorbox}
\usepackage[switch]{lineno}
\usepackage{url}

\title{TelcoAI: Advancing 3GPP Technical Specification Search through Agentic Multi-Modal Retrieval-Augmented Generation}

\author{
    Rahul Ghosh\textsuperscript{1}, Chun-Hao Liu\textsuperscript{1}, Gaurav Rele\textsuperscript{1}, Vidya Sagar Ravipati\textsuperscript{1}, Hazar Aouad\textsuperscript{2}\\
    \textsuperscript{1}Generative AI Innovation Center, Amazon Web Services (AWS)\\
    \textsuperscript{2}Bouygues Telecom\\
    {\tt\small \{rahulgh, chunhaol, grele, ravividy\}@amazon.com}\\
    {\tt\small haouad@bouyguestelecom.fr}
}

\begin{document}
\maketitle
\begin{abstract}
The 3rd Generation Partnership Project (3GPP) produces complex technical specifications essential to global telecommunications, yet their hierarchical structure, dense formatting, and multi-modal content make them difficult to process. While Large Language Models (LLMs) show promise, existing approaches fall short in handling complex queries, visual information, and document interdependencies. We present TelcoAI, an agentic, multi-modal Retrieval-Augmented Generation (RAG) system tailored for 3GPP documentation. TelcoAI introduces section-aware chunking, structured query planning, metadata-guided retrieval, and multi-modal fusion of text and diagrams. Evaluated on multiple benchmarks—including expert-curated queries—our system achieves 87\% recall, 83\% claim recall, and 92\% faithfulness, representing a 16\% improvement over state-of-the-art baselines. These results demonstrate the effectiveness of agentic and multi-modal reasoning in technical document understanding, advancing practical solutions for real-world telecommunications research and engineering.
\end{abstract}

\input{./01_introduction}
\input{./02_relatedWork}
\input{./03_method}
\input{./04_experiment}
\input{./05_results}

\input{./06_conclusion}

\section*{Limitations}
While our agentic multi-modal RAG system advances the state-of-the-art in 3GPP technical specification search, it has several limitations. First, the current implementation is tailored to .docx documents containing both text and embedded diagrams, which aligns with the format of 3GPP specifications. However, other relevant multi-modal sources in the telecommunications domain—such as PDFs with scanned images, Markdown files with external media, and presentation slides—are not yet supported. Expanding multi-modal ingestion to accommodate these diverse formats remains future work. Second, our system is limited to single-turn queries and does not model multi-turn interactions or conversational history, which are important for complex, context-dependent tasks in technical support scenarios. Future work should explore multi-step search agents that can decompose complex queries, maintain conversational state, and iteratively refine searches based on intermediate results. Incorporating agentic workflows—such as an agentic planner for action planning and execution monitoring—could enable sophisticated reasoning chains where the system autonomously decides when to retrieve additional context, cross-reference multiple specifications, or request clarification from users. Extending our framework to handle multi-turn, multi-modal dialogue is a promising direction. Lastly, while we emphasize accuracy and capability in this research prototype, we have not yet optimized for system efficiency, inference latency, or integration overhead, which are critical for real-world deployment. Addressing these engineering aspects is necessary for transitioning our system from research to production.



\appendix

\section{Appendix}
\label{sec:appendix}
\input{./07_appendix}
\end{document}

%% file: 01_introduction.tex
\section{Introduction}
\label{sec:Introduction}

The 3rd Generation Partnership Project (3GPP) \footnote{\url{https://www.3gpp.org}} is responsible for developing and publishing technical specifications for technologies like 3G, 4G, 5G, and beyond, that form the foundation of global mobile telecommunications standards. These specifications are created by various Technical Specification Groups (TSGs) and their associated Working Groups (WGs). The resulting documentation is vast and complex, often spanning hundreds of pages, multiple releases, and containing thousands of equations, tables, figures, and references. This complexity poses significant challenges for engineers and researchers tasked with interpreting, implementing, and staying current with these standards. The emergence of Large Language Models (LLMs) has created new opportunities in the telecommunications domain~\cite{zhou2024largelanguagemodelllm}, particularly in their ability to process and understand technical documentation for network design and operations. With the advent of Generative AI (GenAI) and LLMs, there is an unprecedented opportunity to develop a Question-Answering (QA) system specifically tailored to 3GPP documentation~\cite{huang2025chat3gpp}. Such a system would efficiently process and retrieve relevant information from this extensive corpus, significantly reducing the cognitive load on engineers.

LLMs offer multiple approaches for developing effective QA systems for 3GPP documentation, with ongoing debates centering around the optimal implementation strategy. The two primary methodologies under consideration are Fine-Tuning and Retrieval-Augmented Generation (RAG), each presenting distinct advantages and challenges. Fine-tuning involves adapting pre-trained LLMs specifically to the telecommunications domain~\cite{erak2024leveraging}, potentially offering more specialized and nuanced responses, but faces limitations due to the scarcity of high-quality training data in the technical domain like 3GPP specifications. RAG, on the other hand, represents a more dynamic non-parametric approach by combining LLMs with real-time information retrieval from vectorized 3GPP documentation~\cite{huang2025chat3gpp}, enabling up-to-date and verifiable responses while maintaining the context of the vast technical specifications. Recent research leveraging RAG-enabled AI platforms shows promising results in reducing human cognitive load through comprehensive solutions for 3GPP documentation, spanning from intelligent digital assistance to automated issue detection~\cite{bornea2024telco}, knowledge management in telecommunications~\cite{lin2023ai5g}, and 3GPP-aware agentic actors capable of issue-detection and problem-solving~\cite{roychowdhury2024evaluation}.

Several approaches—such as TSpec-LLM~\cite{nikbakht2024tspec}, Chat3GPP~\cite{huang2025chat3gpp}, Telco-RAG~\cite{bornea2024telco}, Telco-DPR~\cite{saraiva2024telco}, and TelecomGPT~\cite{zou2024telecomgpt}—have explored techniques including domain-specific dataset construction, advanced chunking strategies, hybrid search pipelines, glossary-guided query reformulation, dense passage retrieval, and instruction-tuned LLMs tailored for telecommunications. These methods underscore the importance of specialized data curation, structured retrieval, and domain adaptation to boost performance in technical question answering. However, several key challenges remain. Existing systems often apply flat or generic chunking strategies that fail to leverage the deeply hierarchical and highly structured nature of 3GPP documents, resulting in the loss of crucial contextual cues. Moreover, current approaches largely ignore the multi-modal nature of specifications, neglecting technical diagrams, tables, and visual schematics that are vital for accurate interpretation. From a reasoning perspective, most implementations rely on simple, single-step retrieval patterns and lack agentic capabilities such as structured query decomposition, multi-hop planning, and cross-document information fusion. These limitations are further amplified by the use of evaluation datasets focused on isolated, atomic queries, which do not reflect the complex, multi-faceted information needs of real-world researchers and engineers. Thus there is a need for a more advanced, multi-modal, and agentic framework capable of handling the intricacies of 3GPP documentation.

To address the unique challenges we introduce TelcoAI, an advanced multi-stage RAG system grounded in agentic reasoning and tailored document understanding. Our framework is specifically designed to leverage the hierarchical structure and visual-rich content of 3GPP documents, integrating structural parsing, section-aware chunking, metadata enrichment, and image-text fusion. Through an agentic architecture, the system performs intelligent query decomposition, multi-step planning, and dynamic retrieval refinement to provide accurate, context-aware responses to domain-specific queries. To assess the effectiveness of our approach, we conduct rigorous evaluations across multiple datasets representative of real-world 3GPP use cases. Results demonstrate strong improvements over state-of-the-art baselines for complex, multi-document and version-sensitive queries. Our key contributions are as follows:\\
\textbf{Agentic multi-stage RAG architecture:} We design a modular retrieval and generation pipeline that integrates hierarchical reasoning, query planning, and answer synthesis—mimicking expert decision-making in telecommunications document navigation.\\
\textbf{Multi-modal document understanding:} We build a custom ingestion pipeline that fuses visual (e.g., technical diagrams) and textual content using metadata-aware alignment, enabling comprehensive interpretation of 3GPP specs.\\
\textbf{Section-aware chunking and structured retrieval:} Our method leverages the inherent document hierarchy to preserve semantic coherence across retrieved contexts, improving downstream answer fidelity.\\
\textbf{Agentic query decomposition and fusion:} We introduce a planner that decomposes complex queries into manageable sub-queries and performs answer fusion over diverse content slices to capture nuanced technical insights.\\
\textbf{Comprehensive evaluation:} We conduct extensive experiments on curated and synthetic datasets\footnote{We are working on releasing the datasets as part of the published paper.} tailored to the 3GPP domain, demonstrating superior performance across multiple retrieval and reasoning tasks.\\
Together, these contributions represent a significant step toward domain-specialized, multi-modal language systems for high-stakes technical environments such as telecommunications standards development.

%% file: 02_relatedWork.tex
\section{Related Works}
\label{sec:Related Works}

\textbf{LLMs in 3GPP Domain:} Several works suggest that carefully prepared small models or RAG systems can significantly enhance LLMs for the 3GPP domain. 
In the realm of customized LLMs, researchers have explored various approaches: fine-tuning Microsoft's Phi-2 with LoRA for 3GPP MCQs~\cite{erak2024leveraging}, creating Tele-LLM through full fine-tuning on curated datasets~\cite{maatouk2024tele} , and fine-tuning BERT variants and GPT-2 for 3GPP documents~\cite{bariah2023understanding}. RAG systems have shown particular promise in the 3GPP domain, with \cite{nikbakht2024tspec} demonstrating the potential of naive RAG approaches, while \cite{huang2025chat3gpp} enhanced capabilities through hierarchical chunking and hybrid search. Further advancements include Telco-RAG utilizing technical glossaries and neural routing~\cite{bornea2024telco} and Telco-DPR offering a curated dataset~\cite{saraiva2024telco}. Our approach differs from existing works by emphasizing enhanced retrieval and context processing techniques, leveraging general-purpose LLMs while addressing domain-specific challenges through architectural innovations such as comprehensive query reformulation, planning strategies, and integrated image understanding to handle the multi-modal nature of 3GPP technical specifications.

\textbf{General RAG Architectures:} The RAG paradigm has evolved significantly from Naive RAG's straightforward indexing and retrieval \cite{10.5555/3495724.3496517} to more sophisticated approaches. Advanced RAG \cite{ma-etal-2023-query, zheng2024take} introduced pre-retrieval enhancements like query rewriting and post-retrieval strategies such as re-ranking, improving output quality. Modular RAG further extended these concepts, supporting multiple data modalities \cite{wang2023knowledgptenhancinglargelanguage}, leveraging multi-query mechanisms \cite{Rackauckas_2024}, and incorporating memory modules \cite{10.5555/3666122.3668021} and routing techniques \cite{mu-etal-2024-query}. Recent developments include task adapters for few-shot adaptation \cite{cheng-etal-2023-uprise, dai2022promptagatorfewshotdenseretrieval}, dynamic frameworks to reduce hallucination \cite{ma-etal-2023-query, khattab2022demonstrate}, and iterative methods for enhanced context integration \cite{shao-etal-2023-enhancing, jiang-etal-2023-active, asai2023selfrag}. Our method aligns with recent RAG advancements while addressing unique 3GPP challenges, incorporating Advanced RAG elements in query reformulation, adopting Modular RAG principles for multi-modal handling, echoing iterative retrieval developments through hierarchical chunking and hybrid search, and innovating in document structure preservation for technical content.

%% file: 03_method.tex
\section{Method}
\label{sec:Method}

\begin{figure*}[t]
    \centering
    \includegraphics[width=\textwidth]{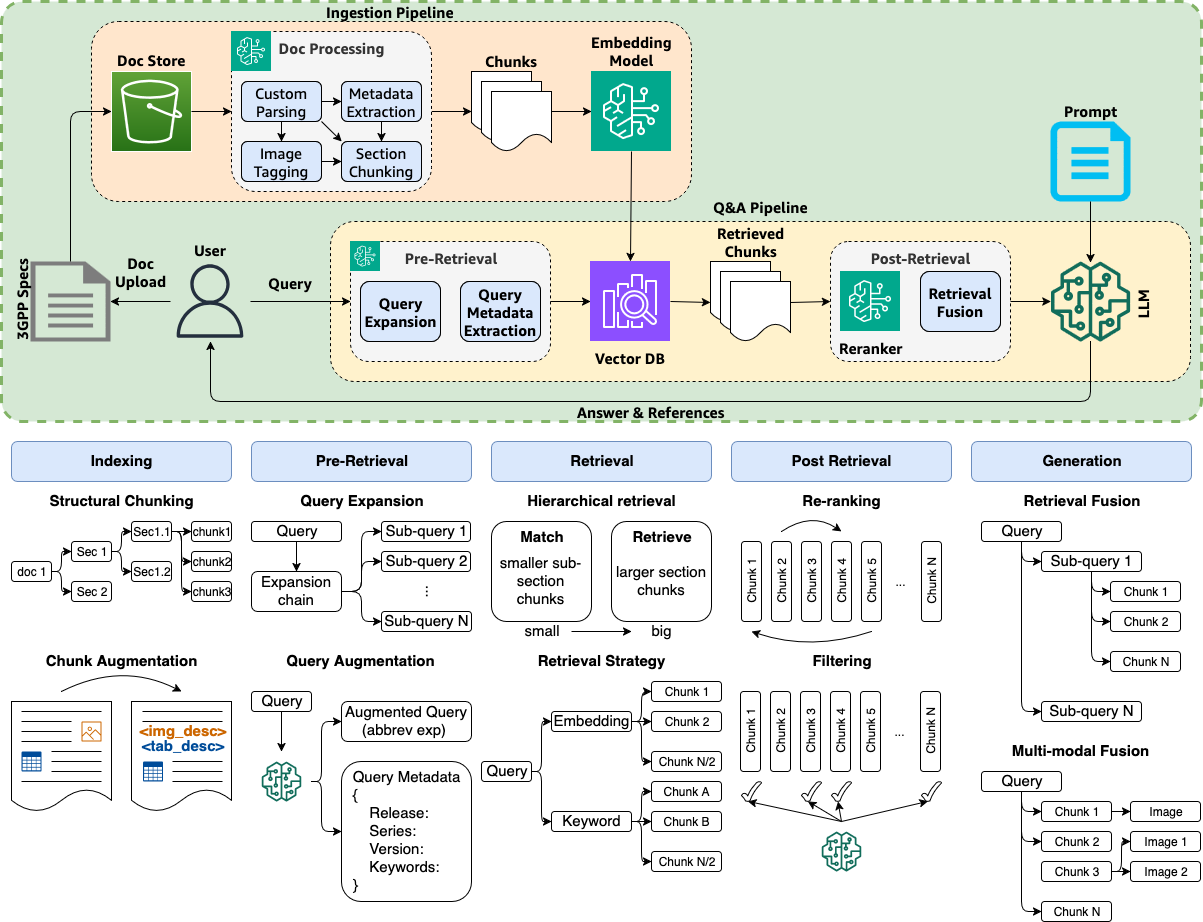}
    \caption{\small \textit{Overview of TelcoAI's architecture showing both ingestion and Q\&A pipelines. Below, detailed workflows show the key components of each stage: indexing, pre-retrieval, context retrieval, post-retrieval and answer generation.}}
    \label{fig:architecture}
\end{figure*}

We present our TelcoAI system architecture (illustrated in Figure~\ref{fig:architecture}) which consists of two main components: (1) Document Ingestion, which processes and indexes the technical specifications, and (2) Question Answering, which handles user queries and generates responses.

\subsection{Document Ingestion}
Document ingestion processes 3GPP technical specifications in Microsoft Word (.docx) format and prepares them for efficient and accurate retrieval. This process involves several key steps: metadata extraction, section-based chunking, table understanding and image tagging.

\subsubsection{Metadata Extraction}

For each document, we extract structured metadata to facilitate targeted retrieval and filtering. Specifically, we extract \textit{Release number} often prefixed with "R" (e.g., "R16"), \textit{Series number} indicating the document series (e.g., "23"), \textit{Specification number} that is the complete identifier (e.g., "23548-i30"), and a three part \textit{Version} (e.g., "V18.3.0"). This metadata is stored alongside each document chunk in the vector database, enabling filtering of retrieval results based on user queries.

\subsubsection{Chunking}
Our system implements an innovative structural chunking approach that preserves the natural organization and semantic coherence of 3GPP technical specifications through section-based decomposition. This approach fundamentally differs from traditional RAG systems' fixed-size or basic hierarchical chunking strategies by maintaining the document's inherent structure. The chunking process is formally defined as a recursive function $\mathcal{C}(d)$ that produces a set of chunks ${c_1, c_2, ..., c_n}$ for a document $d$. The algorithm implements a depth-first traversal through the document's hierarchical structure:
\begin{equation}
c_i = f_{chunk}(d, h_i, h_{i+1}, s_{max}, o),
\end{equation}
where $h_i$ and $h_{i+1}$ represent consecutive section headings, $s_{max}$ is the maximum chunk size, and $o$ is the overlap percentage. The algorithm recursively processes sections and subsections until reaching either: (1) a section that fits within $s_{max}$, or (2) the most granular subsection level.

The chunking process is formally defined as a function $\mathcal{C}(d)$ that produces a set of chunks ${c_1, c_2, ..., c_n}$ for a document $d$, where each chunk corresponds to a meaningful section or subsection in the document's structure. The algorithm implements a chunk creation function:
\begin{equation}
c_i = f_{chunk}(d, h_i, h_{i+1}),
\end{equation}
where $h_i$ and $h_{i+1}$ represent consecutive section headings. To maintain the document's structural integrity, each chunk is mapped to its position in the document hierarchy through function $\mathcal{H}(c_i) = (l_i, p_i)$, where $l_i$ represents the heading level and $p_i$ indicates the position within its parent section. This structured approach ensures that semantically related content remains cohesive, even when technical concepts span multiple paragraphs within a logical section. The preserved hierarchical information is crucial for retrieval and answer generation phases, where understanding the context and relationships between different sections becomes essential. 

\subsubsection{Image Tagging and Integration}
To integrate visual and tabular information from 3GPP technical specifications, TelcoAI implements a comprehensive multi-modal processing strategy. This approach embeds visual and tabular information directly into text chunks through natural language descriptions generated using an LLM (we use Anthropic's Claude 3.5 Sonnet v2~\cite{claude35} in our implementation). During indexing, we transform chunks containing visual elements using:
\begin{equation}
c_i' = f_{tag}(c_i, V_j) = c_i \oplus \text{[DESC]}_{j} \oplus m_j,
\end{equation}
where $c_i'$ is the transformed chunk, $V_j$ represents either an image $I_j$ or table $T_j$, $\text{[DESC]}_{j}$ is the corresponding natural language description, $m_j$ is a unique element marker ([IMG\_XXX] - Simple sequential numbering for each document), and $\oplus$ represents concatenation. The description generation follows:
\begin{equation}
\text{[DESC]}_{j} = \begin{cases}
    f_{img2text}(I_j) & \text{if } V_j \text{ is an image} \\
    f_{tab2text}(T_j) & \text{if } V_j \text{ is a table}
\end{cases}
\end{equation}
This approach preserves semantic relationships between text and visual elements while enabling natural language matching for multi-modal queries. During retrieval and generation, the embedded descriptions participate in semantic matching alongside textual content, while element markers provide access to original visual elements for comprehensive response generation.

\subsection{Question Answering}
Our question answering pipeline consists of four stages: pre-retrieval processing, context retrieval, post-retrieval processing, and answer generation.

\subsubsection{Pre-retrieval}
The pre-retrieval stage prepares user queries for effective retrieval through two main techniques: a) Query Reformulation and Planning and b) Query metadata extraction

\paragraph{Query Reformulation and Planning:} Real-world questions on 3GPP specifications are often complex, requiring synthesis of information across multiple releases, specifications, and sections, as well as iterative exploration to fully address the query intent.

\begin{tcolorbox}[colback=gray!5!white,colframe=gray!75!black,title=Example Complex Query]
\small
"What are the changes to the standard related to MEC interfaces between versions R17 and R18 and between R18 and R19 of specification 23.558? What are the functional additions of these new interfaces?"
\end{tcolorbox}

This complexity motivates our advanced query reformulation and planning strategy, designed to bridge the gap between complex user queries and the structured nature of technical documentation. Given a user query $q$, our query reformulation process leverages an LLM with a planning-aware prompt to develop a question-answering plan and generate a sequence of sub-queries:
\begin{equation}
f_{qr}(q) = \mathcal{M}_{LLM}(p_{qr}, q) = \{q_1, ..., q_m, f_1, ..., f_n\}
\end{equation}
where $q_1$ to $q_m$ ($m \leq 3$) are the core sub-queries that directly address the main question, and $f_1$ to $f_n$ ($m, n \leq 5$) are suggested follow-up queries planned by the LLM to enable deeper exploration of related aspects. Here $p_{qr}$ is a carefully designed query reformulation and planning prompt, and $\mathcal{M}_{LLM}$ is the LLM. Each query is crafted to be standalone, retaining the full context of the original question, including crucial elements such as specification references and release information. This structured approach enables more precise retrieval and allows our system to systematically explore complex technical concepts through planned, iterative questioning.

\paragraph{Query Metadata Extraction:} To leverage the 3GPP technical structure, we extract metadata and expand abbreviations from user queries to enable focused retrieval. Our metadata extraction function processes each query to extract release numbers (e.g., "R16", "R17"), series numbers (e.g., "23"), and specification identifiers (e.g., "23.588.h00"). The extraction process is implemented using an LLM with a carefully designed metadata extraction and abbreviation extraction prompt. For example, from the query "Compare the architecture changes in specification 23.558 between R17 and R18", the system extracts {"release": ["17", "18"], "series": ["23"], "specification": ["23.558"]}, enabling focused retrieval by filtering candidate chunks to those matching the specific releases and specifications mentioned in the query.

\subsubsection{Context Retrieval}
We perform context retrieval from our knowledge base through a multi-stage process that combines dense vector representation and hybrid search strategies. For each sub-query $q_i$ from the reformulation step, we perform context retrieval from our knowledge base through hybrid search, and hierarchical retrieval. First, we encode the query into a dense vector representation, $\mathbf{v}_q = \mathcal{E}(q_i)$. We perform a hierarchical hybrid search combining dense vector similarity and sparse lexical matching:
\begin{equation}
sim(q_i, c) = \alpha \cdot cos(\mathbf{v}_q, \mathbf{v}_c) + (1-\alpha) \cdot lex(q_i, c),
\end{equation}
where $cos(\mathbf{v}_q, \mathbf{v}_c)$ is the cosine similarity between query and chunk vectors, $lex(q_i, c)$ is a lexical similarity function based on BM25~\cite{10.1561/1500000019}, and $\alpha$ is a weighting parameter (set to 0.5 in our experiments). The search operates hierarchically, first retrieving smaller sub-section chunks and then expanding to include their parent sections.

\subsubsection{Post-Retrieval Processing}
The retrieved chunks, initially consisting of the top 40 matches, undergo metadata-based filtering using attributes extracted during pre-retrieval, keeping only chunks with matching metadata. The remaining chunks are then reranked based on similarity scores, with the top 5 selected for the final result, ensuring both content relevance and metadata alignment.

\subsubsection{Answer Generation}
The answer generation phase of our system synthesizes information from multiple retrieved chunks and modalities to produce comprehensive responses to user queries. This process involves three key components: retrieval fusion, multi-modal fusion, and structured response generation.

\paragraph{Retrieval Fusion:} For each sub-query $q_i$, multiple chunks are retrieved, potentially containing complementary or overlapping information. Our retrieval fusion mechanism combines these chunks into a coherent context:
\begin{equation}
\mathcal{C}(q) = \bigcup_{q_i \in f_{qr}(q)} \mathcal{R}(q_i).
\end{equation}
The chunks are ordered by relevance score to prioritize the most salient information. For chunks containing image references, we retrieve the associated images using our image mapping:
\begin{equation}
I_c = {I_j | \exists \mathcal{T}(I_j) = (c, cap_j, ref_j)}.
\end{equation}

\paragraph{Multi-modal Fusion:} Our system performs multi-modal fusion by combining textual and visual information:
\begin{equation}
\mathcal{M}(q_i) = f_{multimodal}(\mathcal{C}(q), {I_1, ..., I_m}),
\end{equation}
where ${I_1, ..., I_m}$ are the images referenced by the markers in the retrieved chunks. This fusion process ensures that technical concepts are explained with appropriate visual support, particularly crucial for architectural descriptions and complex technical diagrams.

\paragraph{Response Generation:} The final response is generated using the Anthropic Claude 3.5 Sonnet v2 model~\cite{claude35}:
\begin{equation}
y = \mathcal{M}_{LLM}(p_{gen}, q, \mathcal{C}(q), I_{\mathcal{C}(q)}),
\end{equation}
where $p_{gen}$ is a carefully designed prompt that instructs the model to analyze the question and context thoroughly, identify relevant information across multiple chunks, incorporate visual elements with appropriate context, perform necessary technical reasoning, and generate a comprehensive answer with explicit references to specifications and sections. For queries requiring multiple steps or comparisons across releases, the system maintains coherence while building a comprehensive answer that addresses all aspects of the original question.

%% file: 04_experiment.tex
\section{Experiments}
\label{sec:Experiments}

\subsection{Datasets}
\label{sec:Datasets}

Our evaluation utilizes four distinct datasets designed for 3GPP technical specifications: Human-QA, Synthetic-QA, Telco-DPR, and TSpec-LLM. \textbf{Human-QA}, our expert-curated dataset, contains 41 technical questions spanning 12 specifications from Releases 16-19, with questions categorized by complexity and requiring single-document retrieval (60\%), cross-document synthesis (25\%), or version comparisons (15\%). \textbf{Synthetic-QA}~\cite{10.5555/3692070.3692736} uses the same document corpus but employs Anthropic Claude 3.5 Sonnet v2~\cite{claude35} with custom prompting to generate multiple-choice questions, validated using Mistral Large 2~\cite{mistral_large_2_2024} and human experts. \textbf{TSpec-LLM}~\cite{nikbakht2024tspec} offers the most comprehensive coverage with 30,137 documents spanning Releases 8-19 (535 million words), preserving original structure and content, and includes an evaluation set of 100 multiple-choice questions categorized by difficulty level.

\subsection{Evaluation Metrics}
For the Human-QA and Synthetic-QA datasets, we evaluate using RAGChecker~\cite{ru2024ragchecker} metrics, which provide both overall performance assessment and detailed diagnostic insights. The evaluation is based on claim-level analysis, where responses and ground-truth answers are decomposed into individual claims for precise comparison. We compute claim-level metrics as follows. Given a model response $m$ and a ground-truth answer $gt$, we first extract their respective claims:
\begin{equation}
Claims_{model} = {c_i^{(m)}}, Claims_{gt} = {c_i^{(gt)}}.
\end{equation}
We then compute recall that evaluates the proportion of ground-truth claims present in the model response:
\begin{equation}
R = \frac{|{c_i^{(gt)} | c_i^{(gt)} \in m}|}{|{c_i^{(gt)}}|}.
\end{equation}
We focus on recall as our primary metric since human-generated ground truth answers tend to be concise and targeted, while model-generated responses often include additional contextual details. In this scenario, recall effectively measures whether the model captures all information from the ground truth, without penalizing the inclusion of relevant supplementary details. For TSpec-LLM, we evaluate accuracy across difficulty levels (Overall, Easy, Intermediate, and Hard) to assess the system's performance on questions of varying complexity.

\subsection{Baselines}
\label{sec:Baselines}
Our evaluation compares our method against several baselines, including base LLMs and specialized RAG systems for 3GPP documentation. \textbf{Base LLMs} like Gemini 1.0~\cite{gemini_1.0_2023}, GPT-4~\cite{gpt4_2023}, Mistral Large 2~\cite{mistral_large_2_2024}, Llama 3.3 70B Instruct~\cite{llama3.3_70b_instruct_2024}, and Anthropic Claude 3.5 Sonnet v2~\cite{claude35}, while demonstrating broad language understanding, lack crucial components for effectively handling 3GPP technical documentation, such as access to the latest specifications, metadata awareness, and multi-modal fusion capabilities. \textbf{Naive RAG} implements the standard retrieve-then-generate pattern without any domain-specific optimizations. \textbf{Telco-RAG}~\cite{bornea2024telco} implements a dual-stage pipeline with query enhancement and retrieval stages. \textbf{Chat3GPP}~\cite{huang2025chat3gpp} uses hierarchical chunking and hybrid retrieval. While these baseline methods were originally evaluated with specific generation models in their respective papers, we extend their evaluation by testing each method with multiple state-of-the-art generation models (Gemini 1.0, GPT-4, Mistral Large 2, Llama 3.3 70B Instruct, and Claude 3.5 Sonnet v2) to ensure a comprehensive comparison across different language models.

\subsection{Implementation Details}
We implement our system using Anthropic Claude 3.5 Sonnet v2~\cite{claude35} as the base LLM and Amazon Titan Text Embeddings v2 (1024 dimensions)~\cite{aws_titan_text_embeddings_v2_ga_2024} for embeddings. For comparison, we evaluate against baseline RAG systems combining two retrievers (BM25 and E5-Mistral~\cite{wang2023improving}) with five generation models (Gemini 1.0, GPT-4, Mistral Large 2, Llama 3.3 70B Instruct, and Claude 3.5 Sonnet v2). Our implementation uses a temperature of 0.7, retrieves top-5 chunks with size of 512 tokens and 20\% overlap. BM25 uses Elasticsearch~\cite{elasticsearch} with default parameters, while dense retrieval uses Titan Text Embeddings v2. All systems use identical chunk size, overlap settings, and retrieval parameters for fair comparison. Evaluation employs RAGChecker with Llama 3.3 70B Instruct as both claim extractor and checker models for detailed performance analysis, providing metrics for overall effectiveness, retrieval quality, and generation reliability.





%% file: 05_results.tex
\section{Results}
\label{sec:Results}

\subsection{Real-world Results}
Table~\ref{tab:realworld} presents the performance comparison of our method against various baselines on the Human-QA and Synthetic-QA datasets, which represent real-world scenarios in querying 3GPP technical specifications. As expected, the results show a clear progression in performance from base models to specialized RAG systems. \textbf{Base Models} demonstrate limited effectiveness in handling 3GPP technical queries, with performance varying significantly across different LLMs. \textbf{Naive RAG} implementations show marked improvements over base models. 
\input{Tables/realworld}
The consistent improvement across both datasets (average improvement of 20-25\% over base models) demonstrates the fundamental value of retrieval augmentation in processing technical specifications. \textbf{Specialized Systems} (Telco-RAG and Chat3GPP) further enhance performance through domain-specific optimizations. Chat3GPP with Claude 3.5 Sonnet v2 achieves the highest baseline performance (Human-QA $R: 0.75$, $CR: 0.73$; Synthetic-QA $R: 0.73$, $CR: 0.72$), representing a significant improvement over naive RAG approaches. \textbf{TelcoAI} achieves the highest performance across both datasets, with consistent improvements over the best baseline. Specifically, our method achieves 16\% and 13.7\% improvement in Recall (0.87 vs 0.75) and Claim Recall (0.83 vs 0.73) respectively in the Human-QA dataset, and 16.4\% and 15.3\% improvement in Recall (0.85 vs 0.73) and Claim Recall (0.87 vs 0.72) respectively in the SyntheticQA dataset.

\subsection{Ablation Studies}
\input{Tables/ablation}

We conducted comprehensive ablation studies on the Human-QA dataset to understand each component's contribution to our system's performance. Starting from a Naive RAG baseline ($R: 0.63$, $CR: 0.6$), we incrementally added components across different pipeline stages, measuring their impact on both Recall ($R$) and Claim Recall ($CR$). Our section-based chunking strategy in the indexing stage provides the first significant improvement ($R: 0.72$, $CR: 0.73$), showing a 14.3\% increase in Recall over the baseline. This demonstrates the importance of preserving document structure and semantic coherence in technical specification processing. Adding query expansion techniques in the pre-retrieval stage further enhances performance ($R: 0.74$, $CR: 0.77$), highlighting the value of comprehensive query understanding.

In the retrieval stage, hierarchical retrieval first improves performance ($R: 0.75$, $CR: 0.78$), followed by additional gains from hybrid retrieval ($R: 0.78$, $CR: 0.79$). Post-retrieval processing, comprising re-ranking and filtering, contributes substantial improvements ($R: 0.82$, $CR: 0.81$) by refining the retrieved context. The final addition of multi-modal fusion in the generation stage yields the highest performance ($R: 0.87$, $CR: 0.83$), representing a total improvement of 38.1\% in Recall and 38.3\% in Claim Recall over the baseline. This significant gain underscores the importance of integrating visual information with textual content in technical documentation understanding. The consistent improvements across both metrics throughout the ablation study validate our system's architecture and the contribution of each component to the overall performance.

To assess the robustness of TelcoAI across different generation models, we extended our evaluation beyond Claude 3.5 Sonnet v2 to include several state-of-the-art LLMs. Table~\ref{tab:model_generalization} presents the performance comparison across four diverse models spanning different architectures and parameter scales. TelcoAI demonstrates robust performance across all evaluated models, with Recall scores ranging from 0.83 to 0.87 on the Human-QA dataset. The variation of only 2-4\% across different generators indicates that our framework's effectiveness is not overly dependent on any specific generation model.

\subsection{Latency Analysis}
\input{Tables/latency}

To evaluate the practical feasibility of our approach, we conducted comprehensive latency benchmarking across all pipeline modules. Table~\ref{tab:latency} presents the performance breakdown for different methods using Claude 3.5 Sonnet v2 as the generation model. TelcoAI achieves a total latency of 11.81s, which represents a reasonable trade-off for the significant accuracy improvements obtained. While this is higher than base LLM (6.73s) and Naive RAG (6.76s), it remains comparable to other specialized systems such as Telco-RAG (9.91s) and Chat3GPP (8.07s). Importantly, the 3-5s additional latency over competing specialized systems translates to substantial accuracy gains: +16\% Recall and +13.7\% Claim Recall over Chat3GPP on the Human-QA dataset.

The latency breakdown reveals several insights. Pre-retrieval processing (2.7s) remains close to Telco-RAG (2.5s), indicating that our query expansion and decomposition techniques add minimal overhead. Retrieval time increases slightly (1.2s vs 0.8-1.0s for baselines) due to our hierarchical and hybrid retrieval mechanisms, but this investment enables more comprehensive context gathering. Post-retrieval processing (0.7s) introduces additional overhead for re-ranking and filtering, but proves essential for context quality improvement as demonstrated in our ablation studies. Generation times remain stable across all methods (6-7s), confirming that our performance gains stem from improved context preparation rather than extended generation processes.

\subsection{Benchmarking Results}
To further validate our system's factual accuracy, we evaluated its performance on the TSpec-LLM benchmark dataset. While this benchmark primarily consists of single-chunk, fact-based multiple-choice questions (simpler than the multi-document, multi-modal queries encountered in real-world scenarios) it provides a valuable test of precise information retrieval. Table \ref{tab:my-table} shows our method achieving state-of-the-art performance with an overall accuracy of 93\%, demonstrating consistent performance across difficulty levels (Easy: 93\%, Intermediate: 93\%, Hard: 92\%).

The results show a clear progression of performance improvements across different approaches. Base models demonstrate limited effectiveness, with accuracies ranging from 30\% (Mistral Large 2) to 67\% (Claude 3.5 Sonnet v2). The introduction of Naive RAG significantly improves performance across all models, with accuracies increasing to 73-82\%. Specialized systems like Telco-RAG and Chat3GPP further enhance performance, with Chat3GPP using Claude 3.5 Sonnet v2 achieving 92\% overall accuracy.

Thus, these results demonstrate that TelcoAI maintains high precision in fact-based scenarios where exact accuracy is crucial. The consistent improvement over strong baselines across all difficulty levels validates the robustness of our approach in both accuracy over fact-based and comprehensiveness in real-world QA.
\input{Tables/benchmark}

%% file: Tables/realworld.tex
\begin{table}[t!]
    \centering
    \caption{\small \textit{Performance comparison (in Recall $R$ and Claim Recall $CR$) across different methods and generation models on Human-QA and Synthetic-QA datasets. Human-QA contains expert-curated complex questions requiring multi-document synthesis, while Synthetic-QA contains LLM generated questions validated by experts. Recall ($R$) measures overall performance while Claim Recall ($CR$) evaluates retriever effectiveness. Base models have no $CR$ values as they operate without retrievers. Within each method category (Naive RAG, Telco-RAG, and Chat3GPP), the retriever remains constant, resulting in a single $CR$ value per category.}}
    \label{tab:realworld}
    \resizebox{\linewidth}{!}{%
        \begin{tabular}{llllll}
            \toprule
            \multirow{2}{*}{Method}     & \multirow{2}{*}{Generation Model} & \multicolumn{2}{l}{Human QA}   & \multicolumn{2}{l}{Synthetic QA} \\
                                        &                                   & $R$ & $CR$          & $R$  & $CR$           \\
            \midrule
            \multirow{5}{*}{\makecell{Base\\Model}} & Gemini 1.0            & 0.34   & \multirow{2}{*}{\Big\uparrow}    & 0.31    & \multirow{2}{*}{\Big\uparrow}                     \\
                                        & GPT 4                             & 0.38   &                                  & 0.35    &                        \\
                                        & Mistral Large 2                   & 0.22   & NA                               & 0.2     & NA                      \\
                                        & Llama 3.3 70B Instruct            & 0.35   & \multirow{2}{*}{\Big\downarrow}  & 0.32    & \multirow{2}{*}{\Big\downarrow}                      \\
                                        & Claude 3.5 Sonnet v2              & 0.42   &                                  & 0.45    &                        \\
            \hdashline
            \multirow{5}{*}{\makecell{Naive\\RAG}}  & Gemini 1.0            & 0.56   & \multirow{2}{*}{\Big\uparrow}    & 0.51    & \multirow{2}{*}{\Big\uparrow}                   \\
                                        & GPT 4                             & 0.61   &                                  & 0.55    &                        \\
                                        & Mistral Large 2                   & 0.54   & 0.6                              & 0.49    & 0.57                      \\
                                        & Llama 3.3 70B Instruct            & 0.6    & \multirow{2}{*}{\Big\downarrow}  & 0.54    & \multirow{2}{*}{\Big\downarrow}                      \\
                                        & Claude 3.5 Sonnet v2              & 0.63   &                                  & 0.57    &                        \\
            \hdashline
            \multirow{5}{*}{\makecell{Telco-\\RAG}}  & Gemini 1.0           & 0.62   & \multirow{2}{*}{\Big\uparrow}    & 0.56    & \multirow{2}{*}{\Big\uparrow}                   \\
                                        & GPT 4                             & 0.65   &                                  & 0.59    &                        \\
                                        & Mistral Large 2                   & 0.64   & 0.67                             & 0.58    & 0.62                      \\
                                        & Llama 3.3 70B Instruct            & 0.68   & \multirow{2}{*}{\Big\downarrow}  & 0.61    & \multirow{2}{*}{\Big\downarrow}                      \\
                                        & Claude 3.5 Sonnet v2              & 0.72   &                                  & 0.65    &                        \\
            \hdashline
            \multirow{5}{*}{\makecell{Chat\\3GPP}}   & Gemini 1.0           & 0.64   & \multirow{2}{*}{\Big\uparrow}    & 0.58    & \multirow{2}{*}{\Big\uparrow}                   \\
                                        & GPT 4                             & 0.71   &                                  & 0.64    &                        \\
                                        & Mistral Large 2                   & 0.68   & 0.73                             & 0.61    & 0.72                      \\
                                        & Llama 3.3 70B Instruct            & 0.72   & \multirow{2}{*}{\Big\downarrow}  & 0.65    & \multirow{2}{*}{\Big\downarrow}                      \\
                                        & Claude 3.5 Sonnet v2              & 0.75   &                                  & 0.73    &                        \\
            \hdashline                            
            TelcoAI                     & Claude 3.5 Sonnet v2              & 0.87   & 0.83                             & 0.85    & 0.83                   \\
            \bottomrule
        \end{tabular}%
    }
\end{table}

%% file: Tables/ablation.tex
\begin{table}[t]
    \caption{\small \textit{Ablation study results (in Recall $R$ and Claim Recall $CR$) on Human-QA dataset showing the incremental performance improvement with the addition of each component. Starting from Naive RAG baseline, each row represents the cumulative effect of adding the corresponding component. The final row shows the overall improvement percentage over the baseline in parentheses.}}
    \label{tab:ablation}
    \resizebox{\linewidth}{!}{
        \begin{tabular}{llll}
            \toprule
            Stage                           & Method                 & R & CR \\
            \midrule
                                            & Baseline (Naive RAG)   & 0.63   & 0.6          \\
            \hdashline
            Indexing                        & Chunking               & 0.72   & 0.73         \\
            \hdashline
            Pre-retrieval                   & Query Expansion        & 0.74   & 0.77         \\
            \hdashline
            \multirow{2}{*}{Retrieval}      & Hierarchical Retrieval & 0.75   & 0.78         \\
                                            & Hybrid Retrieval       & 0.78   & 0.79         \\
            \hdashline
            \multirow{2}{*}{Post Retrieval} & Reranking              & 0.79   & 0.8          \\
                                            & Filtering              & 0.82   & 0.81         \\
            \hdashline
            Generation                      & Multi-modal Fusion      & 0.87 (38.1\%)   & 0.83 (38.3\%)\\
            \bottomrule
        \end{tabular}
    }
\end{table}

\begin{table}[t]
\centering
\small
\caption{Performance of TelcoAI across different generation models on the Human-QA dataset, demonstrating consistent effectiveness independent of the underlying LLM.}
\label{tab:model_generalization}
\begin{tabular}{@{}lc@{}}
\toprule
\textbf{Generation Model} & \textbf{Recall} \\ \midrule
Claude 3.5 Sonnet v2 & \textbf{0.87} \\
Claude 3 Opus & 0.85 \\
Llama 3.3 70B Instruct & 0.84 \\
Mistral Large 2 & 0.83 \\ \bottomrule
\end{tabular}
\end{table}

%% file: Tables/latency.tex
\begin{table}[t]
    \centering
    \small
    \caption{Latency comparison across different methods and pipeline stages. All measurements in seconds, using Claude 3.5 Sonnet v2 as the generation model.}
    \label{tab:latency}
    \resizebox{\linewidth}{!}{
        \begin{tabular}{@{}lccccc@{}}
        \toprule
        \textbf{Method} & \textbf{\begin{tabular}[c]{@{}c@{}}Pre-Ret.\\ (s)\end{tabular}} & \textbf{\begin{tabular}[c]{@{}c@{}}Retrieval\\ (s)\end{tabular}} & \textbf{\begin{tabular}[c]{@{}c@{}}Post-Ret.\\ (s)\end{tabular}} & \textbf{\begin{tabular}[c]{@{}c@{}}Generation\\ (s)\end{tabular}} & \textbf{\begin{tabular}[c]{@{}c@{}}Overall\\ (s)\end{tabular}} \\ \midrule
        Base LLM & - & - & - & 6.73 & 6.73 \\
        Naive RAG & - & 0.5 & - & 6.26 & 6.76 \\
        Telco-RAG & 2.5 & 0.8 & - & 6.61 & 9.91 \\
        Chat3GPP & - & 1.0 & - & 7.07 & 8.07 \\
        {TelcoAI} & {2.7} & {1.2} & {0.7} & {7.21} & {11.81} \\ \bottomrule
        \end{tabular}
    }
\end{table}

%% file: Tables/benchmark.tex
\begin{table}[]
    \caption{\small \textit{Performance (in accuracy \%) comparison on TSpec-LLM benchmark across different methods and generation models. Questions are categorized by difficulty level: Easy, Intermediate (Inter.), and Hard.}}
    \label{tab:my-table}
    \resizebox{\linewidth}{!}{%
        \begin{tabular}{llllll}
            \toprule
            Method                      & Generation Model                  & Overall & Easy & Inter. & Hard \\
            \midrule
            \multirow{5}{*}{\makecell{Base\\Model}} & Gemini 1.0            & 46      & 67   & 37           & 36   \\
                                        & GPT 4                             & 51      & 80   & 47           & 26   \\
                                        & Mistral Large 2                   & 30      & 37   & 26           & 26   \\
                                        & Llama 3.3 70B Instruct            & 47      & 63   & 45           & 26   \\
                                        & Claude 3.5 Sonnet v2              & 67      & 73   & 71           & 47   \\
            \hdashline
            \multirow{5}{*}{\makecell{Naive\\RAG}}  & Gemini 1.0            & 75      & 93   & 65           & 66   \\
                                        & GPT 4                             & 82      & 89   & 85           & 61   \\
                                        & Mistral Large 2                   & 73      & 77   & 75           & 63   \\
                                        & Llama 3.3 70B Instruct            & 81      & 87   & 82           & 68   \\
                                        & Claude 3.5 Sonnet v2              & 82      & 90   & 80           & 74   \\
            \hdashline
            \multirow{5}{*}{\makecell{Telco\\RAG}}  & Gemini 1.0            & 83      & 87   & 87           & 64   \\
                                        & GPT 4                             & 84      & 90   & 87           & 64   \\
                                        & Mistral Large 2                   & 85      & 90   & 84           & 79   \\
                                        & Llama 3.3 70B Instruct            & 85      & 93   & 86           & 79   \\
                                        & Claude 3.5 Sonnet v2              & 91      & 90   & 92           & 90   \\
            \hdashline
            \multirow{5}{*}{\makecell{Chat\\3GPP}}   & Gemini 1.0           & 85      & 90   & 82           & 76   \\
                                        & GPT 4                             & 86      & 92   & 87           & 75   \\
                                        & Mistral Large 2                   & 87      & 88   & 86           & 82   \\
                                        & Llama 3.3 70B Instruct            & 87      & 90   & 84           & 90   \\
                                        & Claude 3.5 Sonnet v2              & 92      & 91   & 92           & 91   \\
            \hdashline
            TelcoAI                     & Claude 3.5 Sonnet v2              & 93      & 93   & 93           & 92   \\
            \bottomrule
        \end{tabular}%
    }
\end{table}

%% file: 06_conclusion.tex
\section{Conclusions}
\label{sec:Conclusions}
We presented TelcoAI, an advanced RAG system specifically designed for processing 3GPP technical specifications, incorporating sophisticated query planning, hierarchical retrieval, and multi-modal fusion capabilities. Our comprehensive evaluation demonstrates significant improvements over existing approaches, achieving 16\% improvement in recall on complex real-world queries and 93\% accuracy on benchmark tasks. The system shows strong performance in both complex scenarios and simpler benchmark tasks, with high faithfulness (0.92) and minimal hallucination (0.05) on multi-document queries. The ablation studies validate our architectural choices, showing meaningful contributions from each component. While opportunities for future work remain, such as handling more diverse visual elements and developing sophisticated cross-version comparison capabilities, our current implementation represents a substantial step forward in making 3GPP technical documentation more accessible and efficiently processable, potentially reducing the cognitive load on engineers working with these complex specifications.

%% file: 07_appendix.tex
\subsection{Tools}

\subsubsection{Query Reformulation}

\begin{tcolorbox}[colback=gray!5!white,colframe=gray!75!black,title=Prompt]
<instruction>
You are a helpful assistant that breaks a user question into multiple consequent sub queries for search, if required.
Write the subqueries so that they are standalone with full context and can be understood with just by themselves.
Do not create new queries with facts other than what is mentioned in the original question
Do not break a query into subqueries and lose the context. do not lose the context of specification and release.
Create subqueries only when necessary. if not needed keep the query as is.
Generate sub queries, one on each line, related to the query given in <query> section.
Do not break a query into subqueries and lose the context. do not lose the context of specification and release.
Be judicious with the number of subqueries generated.
if the user question is simple enough to not be broken into sub queries return the original question.
Do not use your self-knowledge about the domain.
Only state in the sub-queries what is present in the original user question.
Do not go beyond 3 sub queries.
Do not ask questions about specifications.
</instruction>
 
<query>
{query}
</query>
 
<format>
{
    "sub\_queries": [
        'Your search question 1 here.',
        'Your search question 2 here.',
        ...
        'Your search question n here.']
}
</format>
 
Provide your response immediately without any preamble/explain/additional text, enclosed in valid JSON format.
\end{tcolorbox}

\paragraph{Example}:

\begin{tcolorbox}[colback=gray!5!white,colframe=gray!75!black,title=Example Complex Query]
"How has the handling of network slicing evolved in the core network authentication process between R16, R17 and R18 of specification 33.501? What security enhancements were introduced in each version?"
\end{tcolorbox}

\begin{tcolorbox}[colback=gray!5!white,colframe=gray!75!black,title=Generated Query Plan]
Initial Queries:\\
    What changes were made to network slice authentication between R16 and R17 in specification 33.501?\\
    What changes were made to network slice authentication between R17 and R18 in specification 33.501?\\
    What specific security enhancements for network slicing were added in each release?\\
Follow-up Analysis:\\
    Analyze the key security mechanisms introduced in each release\\
    Compare authentication procedures across different slice types\\
    Investigate backward compatibility considerations
\end{tcolorbox}

\subsubsection{Query Metadata Extraction}
\begin{tcolorbox}[colback=gray!5!white,colframe=gray!75!black,title=Prompt]
You are an expert in 3GPP (3rd Generation Partnership Project) Technical Specifications;
these are specifications relevant to mobile communications.
 
<instruction>
- Parse a 3GPP query to extract the three entities contained therein, i.e., release, series, and specification.
- A 'release' (may also be referred to as a 'version') is a number such as 16, 17, ... 19, etc. A release may also be preceded by an 'R', like this: R16, R17, etc.
- A 'series' is typically a number, such as 23.
- A specification is string, typically starting with a series followed by a more digits, with possibly a sub-identifier, like this:  23588, 23.588, or 23.588.h00.
- Hint: if you are confident of the specification (e.g., 23588), then the release is the first two digits, like this: 23.
- If none of the three can be extracted from the queries, return with null.
- Return your answer as JSON with no other commentary.
- Always put the values in a list
- use the following databank if some information is missing
</instruction>
 
<query>
{query}
</query>
\end{tcolorbox}

\subsection{Prompts}
\subsubsection{Answer Generation}
\begin{tcolorbox}[colback=gray!5!white,colframe=gray!75!black,title=Prompt]
You are a 3GPP Question Answering agent specialized in providing accurate and substantiated responses to technical queries based on given context.
Your persona is that of a knowledgeable and meticulous expert who carefully analyzes the question, relevant context, and performs necessary calculations to arrive at a precise and well-reasoned answer.
 
<Task>
Given a user question enclosed in <question> tags and potentially relevant context provided in <context> tags, your task is to provide a complete and correct answer to the question based on the given context.
</Task>
 
<instruction>
To answer the question, think step by step:
1. Read the question carefully and thoroughly understand its requirements.
2. Review the provided context paragraph(s) and identify all information relevant to answering the question.
3. If required, perform any necessary calculations such as sums, divisions, or other operations to derive the answer. Show your work and math clearly.
4. Formulate a final answer that directly responds to the question, grounded in and substantiated by the given context.
5. Present any numerical values in your answer in a rounded and easy-to-read format with appropriate units.
6. Enclose your final answer in an <answer></answer> tag and all the "doc\_name" used to answer the questions in <docs></docs> tag.
7. Include all relevant and exhaustive information from the context to support your answer. Provide a clear explanation justifying your response.
8. Do not answer the question with "Based on the provided context" or anything similar. Just providing answer is enough.
9. If image is not provided, try to answer using the textual context that should contain figure explanation.
</instruction>
 
<question>
{question}
</question>
 
<context>
{context}
</context>
\end{tcolorbox}

\subsection{Detailed Performance Analysis}
\label{sec:Detailed}
\input{Tables/details}

Table~\ref{tab:details} provides comprehensive insights into our method's performance on the Human-QA dataset using RAGChecker metrics. The overall effectiveness metrics show a high Recall (0.87) combined with moderate Precision (0.35), yielding an F1 score of 0.46. Note, that the low precision is choice to prioritize comprehensive information retrieval, often providing additional relevant context beyond the ground truth answers, rather than indicating inaccuracy in the responses.

The retrieval quality metrics demonstrate strong performance in information gathering, with high Claim Recall (0.83) paired with lower Context Precision (0.34). This pattern indicates successful capture of essential information while including broader contextual content from the technical specifications. Such behavior is particularly beneficial in the 3GPP domain, where technical concepts are often interconnected and additional context enhances understanding.

Our system exhibits exceptional reliability in response generation, as evidenced by the generator metrics. Strong Context Utilization (0.71) demonstrates effective use of retrieved information, while minimal Hallucination (0.05) and Self Knowledge (0.03) scores indicate strict grounding in the source documentation. The high Faithfulness score (0.92) further confirms that our generated responses accurately represent the technical specifications. These metrics validate our system's effectiveness in handling complex 3GPP documentation tasks, successfully balancing comprehensive coverage with accurate information representation.

%% file: Tables/details.tex
\begin{table}[]
    \caption{\small \textit{Detailed performance analysis on Human-QA dataset using RAGChecker metrics.}}
    \label{tab:details}
    \resizebox{\linewidth}{!}{%
        \begin{tabular}{lll}
            \toprule
            \multirow{3}{*}{Overall   Metrics}   & Precision           & 0.35 \\
                                                 & Recall              & 0.87 \\
                                                 & F1                  & 0.46 \\
            \hdashline
            \multirow{2}{*}{Retriever   Metrics} & Claim Recall        & 0.83 \\
                                                 & Context Precision   & 0.34 \\
            \hdashline
            \multirow{4}{*}{Generator   Metrics} & Context Utilization & 0.71 \\
                                                 & Hallucination       & 0.05 \\
                                                 & Self Knowledge      & 0.03 \\
                                                 & Faithfulness        & 0.92 \\
            \bottomrule
        \end{tabular}%
    }
\end{table}